\documentclass[10pt,conference,final]{IEEEtran}
\IEEEoverridecommandlockouts
\usepackage{cite}
\usepackage{amsmath,amssymb,amsfonts}
\usepackage{algorithmic}
\usepackage{graphicx}
\usepackage{textcomp}
\usepackage{xcolor}
\usepackage[inline]{enumitem}
\usepackage{caption}
\usepackage{subcaption}
\usepackage{xcolor}
\usepackage{tcolorbox}
\usepackage{listings}
\usepackage{tabularx}
\usepackage{colortbl}
\usepackage{url}
\lstdefinelanguage{ODD}
{
  morekeywords={Include, Exclude, of, for, is,are, not, Conditional, AND, OR},
  morecomment=[l]{\#},
  morestring=[b]"
}

\definecolor{lightgrey}{gray}{0.95}
\definecolor{codegreen}{rgb}{0,0.6,0}
\definecolor{codeblue}{rgb}{0,0,0.7}
\definecolor{codegray}{rgb}{0.5,0.5,0.5}
\definecolor{codepurple}{rgb}{0.58,0,0.82}

\def\BibTeX{{\rm B\kern-.05em{\sc i\kern-.025em b}\kern-.08em
    T\kern-.1667em\lower.7ex\hbox{E}\kern-.125emX}}
\begin{document}

\title{Explaining Unreliable Perception in Automated Driving: A Fuzzy-based Monitoring Approach

\thanks{This work has been partially supported by the Federal Ministry of Education and Research (BMBF) as part of MANNHEIM-AutoDevSafeOps (reference number 01IS22087E). \copyright ~\the\year~Aniket Salvi. All rights reserved. 
This manuscript represents the author's original work. No part of this manuscript may be reproduced, distributed, or transmitted in any form or by any means without the prior written permission of the author, except for brief quotations for academic purposes with proper citation. The author reserves all rights, including the right to use this work in future publications, including the author's PhD thesis. Limited distribution rights are granted to AutoDevSafeOps solely for internal review purposes.}
}

\author{
\IEEEauthorblockN{Aniket Salvi$^{1}$ and Mario Trapp$^{3}$}
\IEEEauthorblockA{\textit{Engineering Resilient Cognitive Systems} \\
\textit{Technical University of Munich}\\
Munich, Germany \\
\{\url{aniket.salvi,mario.trapp}\}\url{@tum.de}}
\and
\IEEEauthorblockN{Gereon Weiss$^{2}$}
\IEEEauthorblockA{\textit{Automation Systems} \\
\textit{Fraunhofer Institute for Cognitive Systems IKS}\\
Munich, Germany \\
\url{gereon.weiss@iks.fraunhofer.de}}
}

\maketitle

\begin{abstract}
Autonomous systems that rely on Machine Learning (ML) utilize online fault tolerance mechanisms, such as runtime monitors, to detect ML prediction errors and maintain safety during operation. However, the lack of human-interpretable explanations for these errors can hinder the creation of strong assurances about the system's safety and reliability. This paper introduces a novel fuzzy-based monitor tailored for ML perception components. It provides human-interpretable explanations about how different operating conditions affect the reliability of perception components and also functions as a runtime safety monitor. We evaluated our proposed monitor using naturalistic driving datasets as part of an automated driving case study. The interpretability of the monitor was evaluated and we identified a set of operating conditions in which the perception component performs reliably. Additionally, we created an assurance case that links unit-level evidence of \textit{correct} ML operation to system-level \textit{safety}. The benchmarking demonstrated that our monitor achieved a better increase in safety (i.e., absence of hazardous situations) while maintaining availability (i.e., ability to perform the mission) compared to state-of-the-art runtime ML monitors in the evaluated dataset.
\end{abstract}


\section{Introduction}
Perception components, which implement tasks such as image classification and object detection using machine learning (ML) models, are vulnerable to changing operating conditions like weather, illumination, and geographical conditions~\cite{bijelicSeeingFogSeeing2020}. These conditions can lead to erroneous predictions, referred to as \textit{misperceptions}, possibly leading to hazardous situations. For instance, a missed front object by a camera-based object detector may result in a collision. It is crucial that these ML-based perception components operate reliably and do not generate \textit{hazardous} misperceptions; such misperceptions require proper mitigation strategies. However, effective mitigation requires knowledge about the cause and impact of these misperceptions on system safety~\cite{cheng2020ac}, which is challenging for \textit{blackbox} ML models (e.g., Deep Neural Networks (DNN)). Runtime monitoring of machine learning models – input deviation monitoring~\cite{cai2020real}, output plausibilization~\cite{inpMon}, and neural network monitoring~\cite{cheng2019runtime} – can detect misperceptions. However, during the design of such monitors, it is essential to consider the evolving nature of ML models due to their re-training, and this must also be taken into account when developing effective mitigation strategies.

Human-interpretable explanations for unreliable ML outputs can facilitate the identification of responsible external factors and targeted mitigation strategies to address their impact on system safety. Interpretability refers to the degree to which a human can understand the cause of a decision made by an ML model~\cite{miller2019explanation}. Interpretable ML techniques provide explanations in two forms: model-agnostic~\cite{ribeiro2016model} or example-based~\cite{aamodt1994case,kim2016examples}. The former separates explanations from the ML model, while the latter provides explanations within training data (e.g., specific images) to explain ML predictions or underlying data distributions. While interpretable ML techniques offer insights into ML behavior, these explanations must be paired with runtime monitors to achieve effective online fault tolerance.

Using data related to ML model behavior (i.e., reliable/unreliable), it is possible to learn suitable explanations linking external factors and their impact in the form of human-readable fuzzy rules. These explanations serve as evidence for reliable ML behavior. This is required for assuring system safety within \textit{suitable} operating conditions (i.e., where the perception component performs reliably). Our proposed technique operates by learning these fuzzy rules from ML model behavior data and then using them as a \textit{predictive monitor}~\cite{bortolussi2021neural} for runtime ML monitoring. The main contributions of this paper are as follows:
\begin{itemize}
    \item an approach for learning a \textit{human-interpretable} monitor for ML-based perception component,
    \item an automated driving case study for evaluating the monitor. 
\end{itemize}

The remainder of this paper is organized as follows: We position our work within state-of-art and provide related work in Section \ref{sec:relWork}. Section \ref{sec:back} provides preliminary background information and motivates our work, outlining the problem statement. Our proposed fuzzy-based monitor and its application for deriving human-interpretable explanations are presented in Section \ref{sec:approach}. In Section \ref{sec:case}, we introduce an automated driving case study that evaluates the fuzzy-based monitor with respect to its interpretability by identifying \textit{suitable} operating conditions for the perception component, supplemented with a safety assurance case for an exemplary driving function. Additionally, we include quantitative benchmarking of its ability to operate as a runtime safety monitor. Finally, Section \ref{sec:conc} concludes this work and outlines future directions.


\section{Related Work}\label{sec:relWork}
ML models' inherent non-deterministic and black-box nature has driven research into various interpretability techniques, enabling humans to understand the reasoning behind ML predictions. One approach involves training \textit{surrogate} interpretable models (e.g., decision trees, linear/logistic regression) alongside ML models to approximate their predictions at both global and local levels~\cite{lundberg2017unified}. Unlike global surrogate models, local surrogate models like LIME~\cite{ribeiro2016should} are trained to approximate an individual ML model prediction. However, these \textit{simple} surrogate models can only \textit{approximate} complex ML models and may not \textit{fully} reflect the ML model predictions. Furthermore, these techniques require identifying interesting sample(s) and training local surrogate models to obtain explanations, making them non-scalable.

Rather than approximating ML model predictions, authors in \cite{langford2021know,langford2023modalas,langfordModularComposableApproach2022} propose learning to approximate ML model behavior\footnote{model's ability to provide \textit{correct} predictions} (e.g., reliable, degraded, unreliable). 
They introduce the concept of a behavior oracle, which is learned from data comprising simulated environmental conditions impacting training inputs (e.g., blurred camera images) and observing ML model behavior on them (e.g., increase in proportion of misperceptions). Trained behavior oracles can then be used during operation to utilize ML model outputs only during environmental conditions deemed trustworthy (e.g., predicted as \textit{reliable} ML model behavior). Similarly, by leveraging simulation environments, \cite{bortolussi2021neural} propose to learn an \textit{predictor} serving as a runtime \textit{predictive monitor} for detecting incorrect ML predictions. However, these approaches still rely on another black-box ML model behavior oracle, leaving the interpretability problem unresolved.

Approaches in \cite{cheng2019runtime} propose monitoring neural network layers to detect erroneous neuron activations indicative of possible misperceptions. Runtime monitors proposed by \cite{geissler2023low} provide human-interpretable neuron activations. However, these approaches require a \textit{white-box} assumption of the ML model and do not account for the evolution of the ML model owing to re-training.

In contrast, our proposed fuzzy-based monitor shares similarities with behavior oracles but can be trained incrementally, yielding human-interpretable explanations for its detections. Additionally, the transparent structure of the learned fuzzy rules enables gathering evidence for ML behavior and using a fuzzy inference system as a runtime safety monitor.
\section{Background}\label{sec:back}
ML models are invariably impacted by small changes in their input resulting from their operation in diverse operating conditions. We differentiate between the impact of external environmental factors on perception component reliability and the safety of systems utilizing these perception components; in the former case, such factors are referred to as Perception-Only (PO) conditions. 

\subsection{Motivation}
Sensing systems used for perception tasks are particularly susceptible to variations in environmental operating conditions under which they operate~\cite{salvi2023adaptively}. These external factors can lead to fluctuations in ML model predictions and their associated confidence scores, resulting in overconfident \textit{incorrect} predictions or underconfident \textit{correct} predictions. This can lead to hazardous situations when ML models produce highly confident erroneous predictions. At the same time, reduced availability of autonomous systems occurs when ML models are overly cautious due to low confidence. To ensure the reliable operation of perception components and produce accurate predictions with high-confidence scores, constituent ML models must undergo additional training with diverse operational context data through MLOps~\cite{guissouma2023continuous,fayollas2020safeops} iterations that occur throughout their lifecycle. Developing \textit{assurance cases}~\cite{cheng2020ac} for the safe operation of autonomous systems requires identifying Perception-Only (PO) conditions, along with statistical evidence, where perception components operate reliably. Despite re-training the ML models, perception components may still produce misperceptions that potentially lead to hazards. Runtime monitors can complement design-time measures by prohibiting system use in untrusted PO conditions.

\subsection{Problem Statement}\label{sub:prob}
For an environmental state, we define its observation $\Omega:=(o,x)$ as a tuple consisting of PO conditions $\mathbf{o}$ and the input recorded by sensor $x \in \mathcal{X}$. A perception component implements a perception task $T:\mathcal{X}\rightarrow\mathcal{Y}$, producing an output $\hat y=T(x)$. For a component $C$ implementing the perception task, an instance $(x,y)$ is a misperception if $y \neq C(x)$. At design-time, given $y$, we aim to learn an interpretable classifier function $\Xi:\Omega\rightarrow \varphi, \varphi \in \{0,1\}$, where $\varphi=1$ refers to the misperception, and 0 otherwise. 
The resulting interpretability must allow us to identify a set of \textit{suitable} PO conditions, $\mathcal{O}=\{\mathbf{o}\}$, in which misperceptions result in \textit{acceptable} proportion of hazards (depicted as $\gamma_C$) such that $\forall \mathbf{o}\in\mathcal{O}, P(hazard|\Omega) \leq \gamma_C$), and can be captured in an assurance case. 

\subsection{Evolving Fuzzy Systems}
Online inference models~\cite{fontenla2013online,skmultiflow} can learn from data streams and provide predictions in their current trained state. Evolving Fuzzy Systems~\cite{lughofer2011evolving} (EFS) can learn Takagi-Sugeno-Kang~\cite{zhang2024takagi} fuzzy rules of the form: "IF $input$ is $A$ THEN $output$". AlMMo~\cite{angelov2017autonomous}, an EFS variant, represents the antecedent A using Empirical Fuzzy Sets~\cite{angelovEmpiricalFuzzySets2018}. The antecedent A (cf.~\eqref{eq:fuzzRule}), a datacloud, has an associated representative prototype, similar to example-based interpretable ML techniques. A \textit{datacloud} is a fuzzy representation of its comprised data instances, similar to a cluster. With AlMMo, it is possible to learn both the dataclouds and fuzzy rules that compose a fuzzy-based inference system from training data. Domain-expert knowledge can be integrated into learning using user-provided prototypes at the beginning of EFS training.

\section{Proposed Approach}\label{sec:approach}
In this section, we present our approach for learning a human-interpretable fuzzy-based monitor that can be used for identifying \textit{suitable} PO conditions. In addition, it can be employed as a runtime safety monitor for ML-based perception components.

\subsection{Overview}
\begin{figure*}[htp]
\centering
\includegraphics[width=0.7\textwidth]{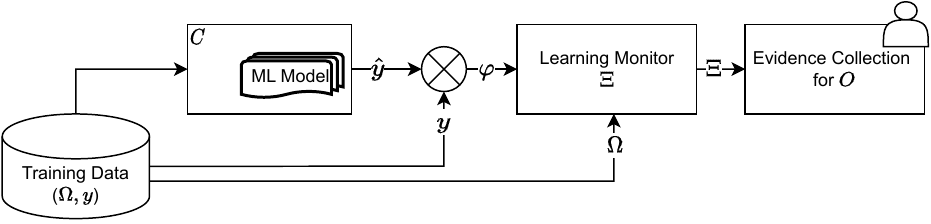}
\caption{Overview of Proposed Fuzzy-based Monitor Learning}
\label{fig:overview}	
\end{figure*}

An overview of our proposal is presented in Fig. \ref{fig:overview}. It involves learning a fuzzy inference system, specifically a fuzzy-based classifier ($\Xi$), used as a runtime predictive monitor $m_C$ for the perception component $C$.
The \textit{Training Data} consists of \textit{labeled} input data - $(x,y)$ (e.g., RGB camera images with 2D bounding boxes) and information about PO conditions - $\mathbf{o}$ (e.g., weather annotations, sensor quality attributes-noise, etc.), under which it was captured. 

During the learning stage, as discussed in Section \ref{subsec:learncloud}, the perception component is evaluated with training data $(x, y)$ to obtain $\varphi$ (cf. \ref{sub:prob}) for a specific instance. Additionally, platform-specific dependencies that impact perception reliability are captured as further inputs~\cite{burton2019confidence,Weiss.2018} (e.g., image blur, etc.). We extend the AlMMo algorithm for learning a fuzzy-based classifier $\Xi$; our extensions are discussed in section \ref{subsec:learncloud}. Once $\Xi$ has attained sufficient prediction accuracy, i.e., can accurately predict $\hat{\varphi}$ for given conditions $\mathbf{o}$, the learned fuzzy rules and transparent datacloud ($\mathcal{D}$) structure are used to identify suitable PO conditions $\mathcal{O}$ and gather evidence for an assurance case (cf.~Section \ref{subsec:specpo}). 

Furthermore, the learned fuzzy rules create a fuzzy inference system that can be used as a predictive runtime monitor ($m_C$), cf. section \ref{subsec:mon}. This runtime monitor can be deployed alongside the perception component $C$ in PO conditions prescribed by $\mathcal{O}$, see Fig.~\ref{fig:explain}, green region with solid \textit{known} boundary. 
The predictive monitor $m_C$ detects unreliable PO conditions (as seen in the red region with dashed \textit{unknown} boundaries) arising from specification uncertainty in $\mathcal{O}$.

The \textit{Training Data} for learning the monitor need not overlap with the data for ML model re-training and may include validation or testing datasets. The learning process can be repeated using an existing learned fuzzy monitor $\Xi$ whenever a re-trained ML model or new training data is available, thanks to AlMMo's online learning capability.

\subsection{Learning Dataclouds for PO Conditions}\label{subsec:learncloud}
Data-driven learning of a fuzzy-based classifier $\Xi$ is possible from scratch using AlMMo's algorithm. The primary objective of this training is to categorize the PO conditions reflected in $\Omega$ based on their contribution to the value of $\varphi$. Essentially, the learned fuzzy rules provide a mapping of these sets of PO conditions (based on their corresponding dataclouds) to misperceptions by ML models. 
To represent the monitorable space of PO conditions, we employ a vector feature-space denoted by $\mathbf{o} = [o_1, o_2, \dots, o_j]$, where $o_j$ represents a measurable operating condition constituting a PO condition. These PO conditions can be measured as categorical literals (e.g., weather and scene annotations) or numerical values (e.g., precipitation intensity [mm/hr], visibility [m]), without requiring the assumption of \textit{independence and identical distribution} (iid). 

\begin{figure}[htbp]
	\centering
	\includegraphics[width=\columnwidth]{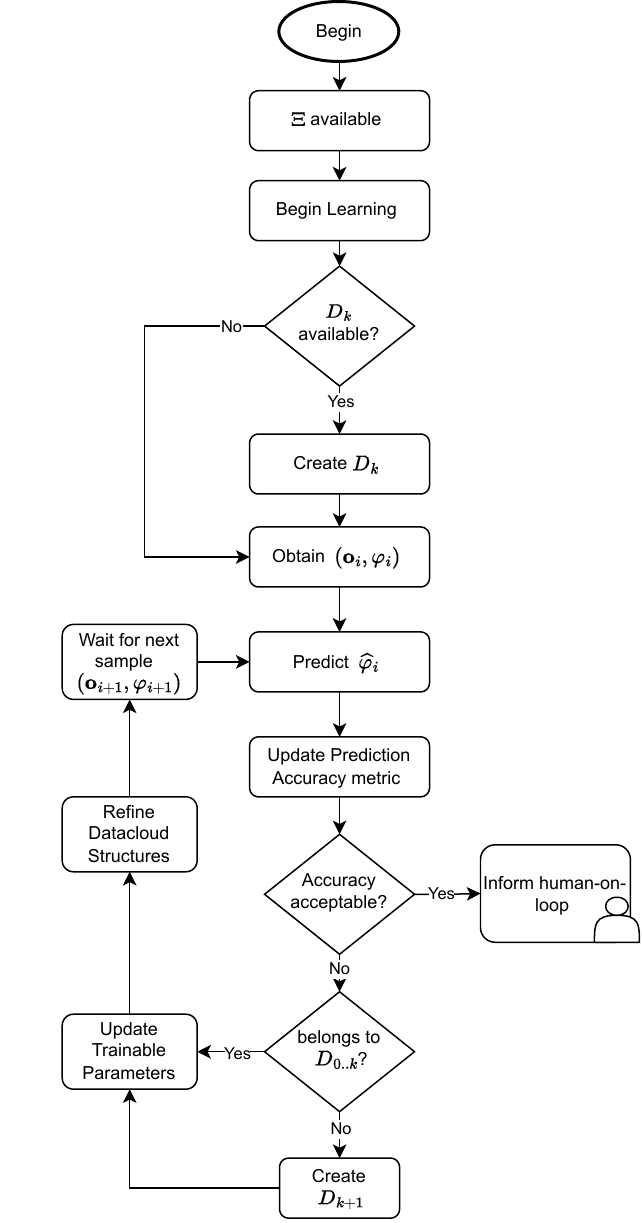}
	\caption{Learning Monitor: Datacloud Discovery}
	\label{fig:proc}	
\end{figure}

The main steps for learning $\Xi$ are outlined in a flowchart in Fig.~\ref{fig:proc}. 
The $\Xi$ training can start using a user-provided PO condition for initializing a datacloud ($\mathcal{D}_k$), referred to as a prototype $\mathbf{p}_k=[p_1,p_2,\dots,p_j]$ (cf. {\fontfamily{pcr}\selectfont
$D_k$ available?}). However, prototypes can also be discovered as part of the learning process. For $\mathbf{p}_k$, a new datacloud $\mathcal{D}_k$ (cf. {\fontfamily{pcr}\selectfont
Create $D_k$}) is created with membership function denoted by unimodal discrete density as - \begin{equation}\label{eq:fuzzmem}
	\mu_k(\mathbf{o})=\frac{1}{1+\frac{{||\mathbf{o}-\mathbf{p}_k||}^2}{\sigma_k^2}}
	\end{equation}
where $\sigma^2=X_k-{||\mathbf{p}_k||}^2$; $\mathbf{p}_k$ and $X_k$ are, respectively, the mean and average scalar product of the dataset comprising PO conditions belonging ($\sim$) to datacloud $\mathcal{D}_k$. The corresponding learned fuzzy rule is represented as -  \begin{equation}\label{eq:fuzzRule}
\text{IF }\mathbf{o} \sim \mathbf{p}_k\text{ THEN }\varphi_k=[1,o_1,o_2,\dots o_j]\mathbf{a}_k
\end{equation}where, $\mathbf{a}_k$ represents the learned consequent for datacloud $\mathcal{D}_k$. 

We assume the \textit{Training Data} may become available as a single instance or batches throughout the lifecycle of the perception component, inline with MLOps. Therefore, the $\Xi$'s prediction accuracy may improve over time with additional training. Upon obtaining $i^{th}$ training instance $(\mathbf{o}_i,\varphi_i)$, we employ a \textit{test-then-train}~\cite{blum1999beating,grzenda2020delayed} strategy to track $\Xi$'s prediction accuracy. 
This involves first predicting $\hat{\varphi}_i$ for input $\mathbf{o}_i$, then updating the rolling \textit{Accuracy} in prediction, and finally using $(\mathbf{o}_i,\varphi_i)$ for training. If $\Xi$ exhibits acceptable\footnote{user-defined based on the application of perception component} prediction accuracy, its training state is saved, and a human-on-loop is notified to review it for evidence collection (cf. {\fontfamily{pcr}\selectfont
Inform human-on-loop}). Otherwise, the training instance is checked for belonging to existing dataclouds within $\Xi$ (cf. {\fontfamily{pcr}\selectfont
belongs to $D_{0..k}$?}). In case of a match, the corresponding datacloud's structure (membership function, prototype, support, etc.) is updated as part of the training process (cf. {\fontfamily{pcr}\selectfont
Update trainable parameters}). 

The support $supp(\mathcal{D}_k)=\{\mathbf{o}|\mu_k(\mathbf{o})>0\}$ represents the set of PO conditions which can be categorized within a datacloud $\mathcal{D}_k$. During the datacloud structure update, the values of $\mathbf{p}_k$ and $supp(\mathcal{D}_k)$ are recursively updated as part of trainable parameters. This updating of trainable parameters ($\mu_k,\mathbf{a}_k$) is carried out as described in \cite{angelov2017autonomous}. If no \textit{belonging} datacloud is found (i.e., no nearest existing $\mathbf{p}_{0\dots k}$), a new datacloud $\mathcal{D}_{k+1}$ is created based on this training instance (cf. {\fontfamily{pcr}\selectfont
Create $D_{k+1}$}).

The existing dataclouds $\mathcal{D}_{0\dots k+1}$ are continuously evaluated according to their contribution to predictions $\hat{\varphi}$ and may be merged or pruned accordingly (cf. {\fontfamily{pcr}\selectfont
Refine Datacloud Structures}). The learning process resumes with the next incoming training sample $(\mathbf{o}_{i+1},\varphi_{i+1})$. 


\subsection{Interpreting PO Conditions}\label{subsec:specpo}
The saved state of $\Xi$ comprises user-defined or automatically discovered dataclouds and their mapping to the value of $\varphi$, contained in fuzzy rules. Each datacloud $\mathcal{D}_k$ is \textit{transparent} and can be represented by its prototype $\mathbf{p}_k$ and associated membership function $\mu_k$. We utilize these dataclouds and associated information to specify suitable PO conditions (i.e., $\mathcal{O}$) and gather evidence for the creation of a safety assurance case (see section \ref{sub:prob}).

\subsubsection{Prototype as PO Condition Representation}\label{subsub:protopo}
We propose to employ the automatically discovered prototypes to specify the PO conditions. Consider an example discovered prototype: $\mathbf{p}_0=[clear, city street, daytime, 0.67]$ representing weather, scene, time-of-day, and image contrast measurable operating conditions $o_j$; mapped to a $\varphi=0$ score by its corresponding fuzzy rule. For a camera-based perception task, an image corresponding to $\mathbf{p}_0$ serves as a human-interpretable example of PO conditions represented by datacloud $\mathcal{D}_0$. Alternatively, a human domain expert may provide a camera image as an input prototype (e.g., used to extract $\mathbf{p}_1$) which will be used in the learning process to create a datacloud $\mathcal{D}_1$. 

\begin{figure}[htbp]
    \centering
    \includegraphics[width=\columnwidth]{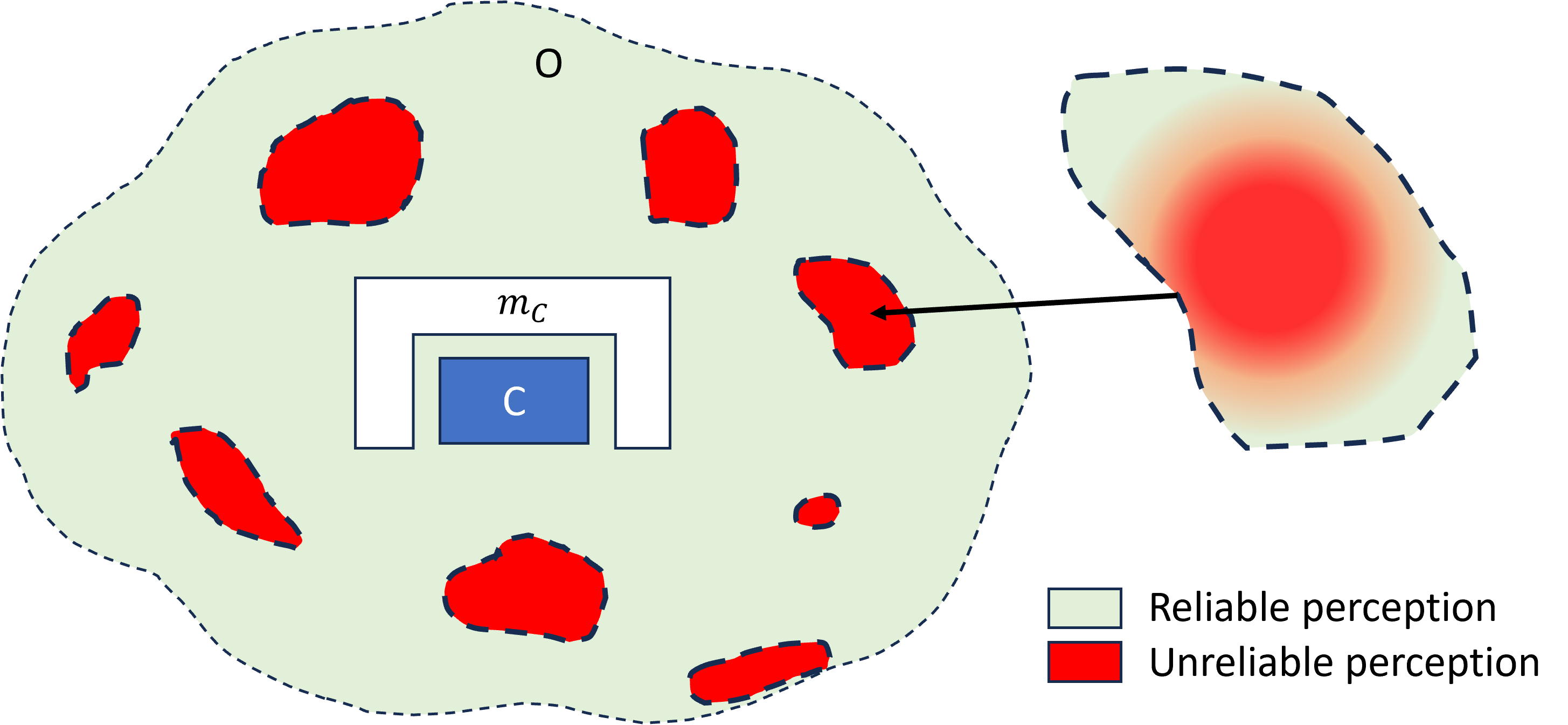}
    \caption{Representing PO conditions with dataclouds}
    \label{fig:explain}
\end{figure}

The prototype $\mathbf{p}_k$ serves as an example of \textit{trusted} or \textit{untrusted} PO conditions that result in 0 or 1 value of $\hat\varphi$, respectively. In our example, PO conditions represented by $\mathcal{D}_0$ could be considered trusted conditions since there are no predicted misperceptions. Our case study, in Section \ref{sec:case}, provides a detailed discussion along with a human-readable specification for $\mathcal{O}$ created from the \textit{shortlisted} $\mathcal{D}_k$'s (i.e., with evidence for reliable operation). Fig.~\ref{fig:explain} shows examples of reliable (\textit{red}) and unreliable (\textit{green}) dataclouds.

\subsubsection{Collecting Evidences for Perception Component}\label{subsub:evidence}
The learned dataclouds and associated information contribute as statistical evidence regarding the reliable operation of the perception component (cf. Fig~\ref{fig:overview}). For example, the probability of misperception (mP) within instances belonging to a datacloud constitutes evidence of perception components reliability and is computed as - \begin{equation*}\label{eq:propmP}
P(mP|\mathcal{D}_k)=\frac{1}{|supp(\mathcal{D}_k)|}\sum_{\mathcal{D}_k}1[\varphi==1]
\end{equation*} where the numerator denotes the count of instances within datacloud with misperception, i.e.,  $\varphi=1$. The evidence obtained for perception component reliability (e.g., $P(mP|\mathcal{D}_k)$) must be linked with perception component safety (cf. section \ref{sub:prob}, $\gamma_C$), and has been discussed in detail with a case study in section \ref{sec:case}. 

Using a single sampled instance like $\mathbf{p}_0$ to represent a specific PO condition feature space poses two challenges. First, the datacloud ($\mathcal{D}_0$) represented by $p_0$ is only a sample and may not fully represent the target feature space. Second, $\mathcal{D}_0$ may include untrusted instances (i.e., $\varphi \neq \hat\varphi$), such as those with a very low degree of membership to $\mathcal{D}_0$ ($\mu_0 << 1$). See Fig.~\ref{fig:explain} for a fuzzy representation of untrusted (red) datacloud. For the former case, we account for a sampling error ($\Delta_q$) in our calculations for creating statistical evidence, cf. \eqref{eq:propmP}. For an evidence related metric $\gamma$ computed for datacloud $\mathcal{D}_k$, the upper bound $\Delta_q$ of q\% confidence error is calculated as - \begin{equation}\label{eq:sampE}
	\Delta_q = v(q)\sqrt{\frac{\gamma(1-\gamma)}{|supp(\mathcal{D}_k)|}}
\end{equation} 
where, $v(q)=\frac{100+q}{200}$ is the quantile of standard normal distribution. For the latter case (i.e., $\mu << 1$), we propose to utilize a runtime monitor (cf. Fig.~\ref{fig:explain}) that can accurately predict if the PO condition is (un)trustworthy, discussed in section \ref{subsec:mon}.

\subsection{Creating Runtime Monitor}\label{subsec:mon}
To complement design-time safety and tackle the challenges discussed in the previous section, a runtime monitor is used to safeguard the perception component, see Fig.~\ref{fig:explain}. Runtime monitoring essentially serves as an online fault tolerance mechanism detecting \textit{untrusthworthy} PO conditions possibly included in suitable dataclouds. This is because $\Xi$'s classification accuracy is often less than 100\% (cf. section \ref{subsec:learncloud}) or due to insufficient consideration of the sampling error $\Delta_q$, cf.~\eqref{eq:sampE}.  
We propose to use $\Xi$ as a runtime predictive monitor $m_C$, monitoring the perception component $C$, collectively referred to as $(C,m_C)$. The runtime monitor $m_C$ is evaluated with a case study, and the benchmarking results have been discussed in section \ref{subsec:runEval}.

\bigskip
In this section, we presented our approach for learning human-interpretable fuzzy monitor, comprising data clouds and fuzzy rules. Using this fuzzy monitor for a perception component was discussed to derive suitable PO conditions (supported by human-verifiable evidence) at design-time and its role as a runtime safety monitor.

\section{Case Study: Automated Driving}\label{sec:case}
We consider an Automated Driving System (ADS) as a case study to demonstrate our proposed approach. ADS utilizes ML-based perception components to partially or fully automate certain driving tasks. However, ADS features enabling driving automation, e.g., Traffic-Jam Assist, Highway Pilot, etc., must be deployed within a specified set of operating conditions, referred to as Operational Design Domain (ODD)~\cite{iso34503:2023}.  

In this section, we evaluate the human interpretability of learned fuzzy-monitor and its role in supporting human-driven activities such as safety assurance case creation for the perception component (cf. section \ref{subsec:specpo}). The main objective of this study is to derive a human-readable ODD specification corresponding to \textit{suitable} PO conditions observed within a training dataset, wherein an exemplary ML-based object-detector (i.e., perception component) operates reliably. Furthermore, this ODD specification must be supplemented with a safety assurance case, and the perception component must be monitored at runtime, refer section \ref{sub:prob}.   

Firstly, we introduce the use-case scenario used within this case-study in section~\ref{sec:UC}. We describe our experimental setup for realizing the use-case scenario in section \ref{sub:expSet}. The experimental results concerning learning of the fuzzy monitor have been presented and discussed in section \ref{sec:fuzzLearn}. Finally, we benchmark our learned fuzzy monitor against several other interpretable (e.g., decision-tree) and non-interpretable (e.g., neural network-based) classifiers used as \textit{predictive} runtime safety monitors in section \ref{subsec:runEval}.  

\subsection{Use-case Scenario}\label{sec:UC}
We consider the use-case scenario \textit{Stopped Car Ahead} (SCA) described in \cite{salay2022missing} and utilize their \textit{Integration Safety Case for Perception} (ISCaP) template to create a safety assurance case. It includes the evidence for the reliable operation of the ML-based 2D object-detector within the specified ODD, in our case $\mathcal{O}$. The characteristics of the integrated perception component are described as follows:
\begin{tcolorbox}[colback=gray!20, colframe=gray!50] 
\begin{center}
	The output of the object-detector is fed to a \textit{Tracker}, which maintains the past states of detected objects. The \textit{Tracker} loses track of a detected object after \textit{9} consecutive missed detections by the object detector,  i.e., 9 False Negatives (FNs). Furthermore, based on physics-based calculation, additional \textit{5} FNs lead to a crash with SCA.
\end{center}
\end{tcolorbox}
We limit the focus of this work only to missed object detections, i.e. FNs. In our case, the misperception (mP) by the object-detector is considered hazardous (i.e., HmP) if we observe an FN within the Region of Interest, see Fig.~\ref{fig:scE}, \textit{blue} bounding-box. Here, the missed car (depicted by a \textit{red} bounding-box) by a 2D object detector can be considered as HmP.

\subsection{Experimental Setup}\label{sub:expSet}

We use the naturalistic driving dataset $\{(x_i,\mathbf{o}_i)\}_n$ for this case-study. The BDD100k~\cite{seita2018bdd100k} evaluation dataset comprises camera images ($x$) captured under diverse PO conditions ($\mathbf{o}$), with $n=10.000$. In addition to available \textit{categorical} values for operating condition attributes (i.e., \textit{weather} ($o_1$), \textit{scene} ($o_2$), \textit{timeofday} ($o_3$)), we capture platform-dependent attributes: \textit{numeric} - (\textit{brightness} ($o_6$), \textit{clearness\_score} ($o_7$), \textit{contrast\_score} ($o_8$)) and \textit{boolean} - (\textit{blurry} ($o_4$), \textit{low\_contrast} ($o_5$)). As an exemplary object-detector, we use the pre-trained and publicly available ML model \textit{yolov5x6}~\cite{yolov5}  as a perception component. In this study, we focus on changing weather conditions, as they have the most significant impact on the defined perception reliability. We perform a 70:30 split of the BDD100k dataset into training  and validation datasets comprising 6127 and 2626 samples, respectively\footnote{samples with \textit{undefined} annotations are removed}. 

The behavior of the exemplary 2D object detector was observed on the training and validation split and recorded as training data $d_{train}$ and $d_{val}$ comprising $(\mathbf{o},\varphi)$ for monitor learning, respectively. An example instance referring to a yolov5x6 misperception from this training data consists of training input: $\mathbf{o}=[clear,highway,daytime,False,False,18.16,0.12,2.18]$, and training output: $\varphi=1$.

\begin{figure*}[htbp]
	\centering
	
	\begin{subfigure}{.31\textwidth}
		\centering
		\includegraphics[width=\linewidth]{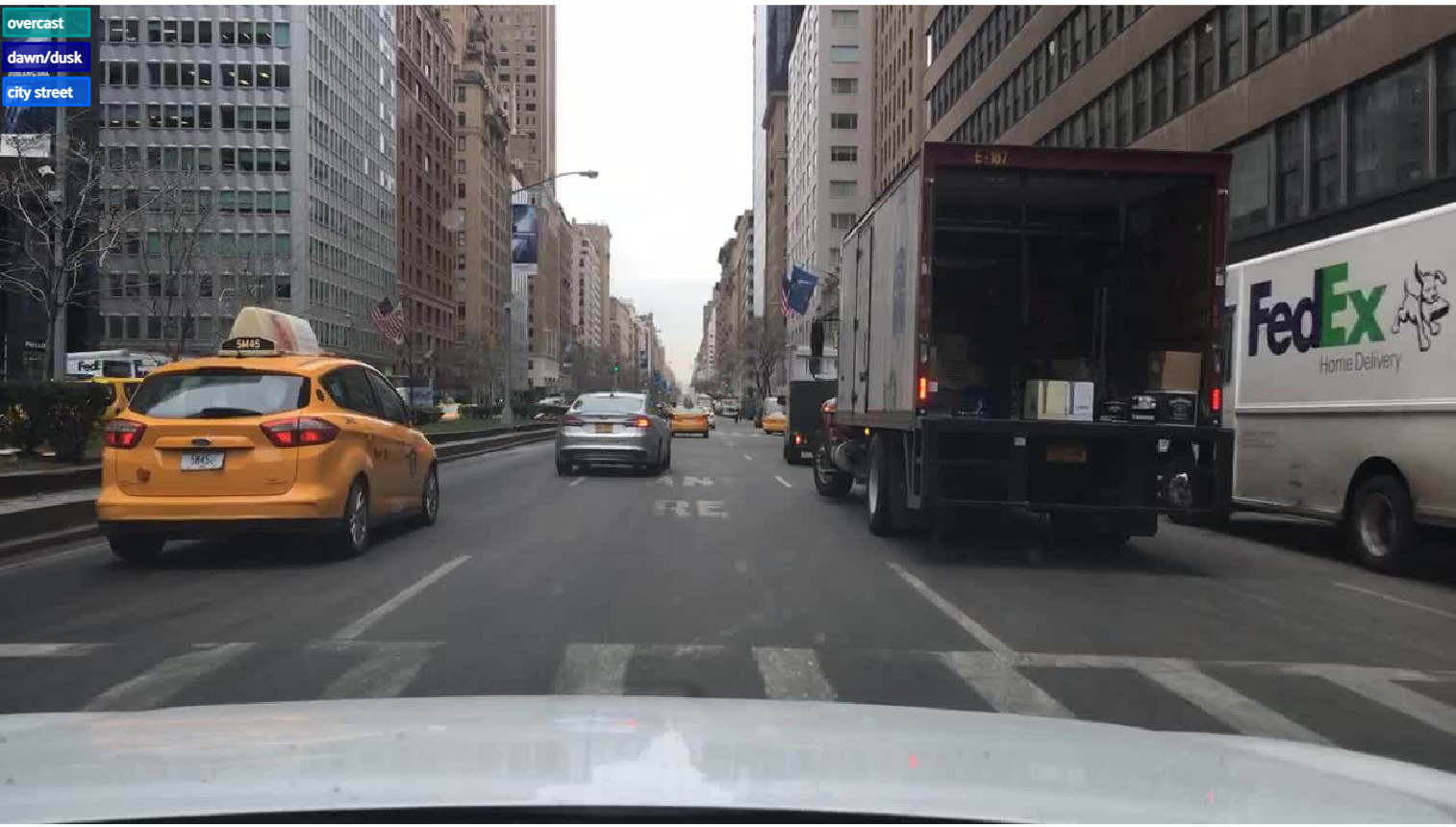}
		\caption{\textcolor{green}{$\mathcal{D}_2$}: overcast,city-street,dawn/dusk}
		\label{fig:d0cloud}
	\end{subfigure}%
	\hfill
	\begin{subfigure}{.31\textwidth}
		\centering
		\includegraphics[width=\linewidth]{figures/accepted/c6.png}
		\caption{\textcolor{green}{$\mathcal{D}_6$}: clear,city-street,daytime}
		\label{fig:d1cloud}
	\end{subfigure}%
	\hfill
	\begin{subfigure}{.31\textwidth}
		\centering
		\includegraphics[width=\linewidth]{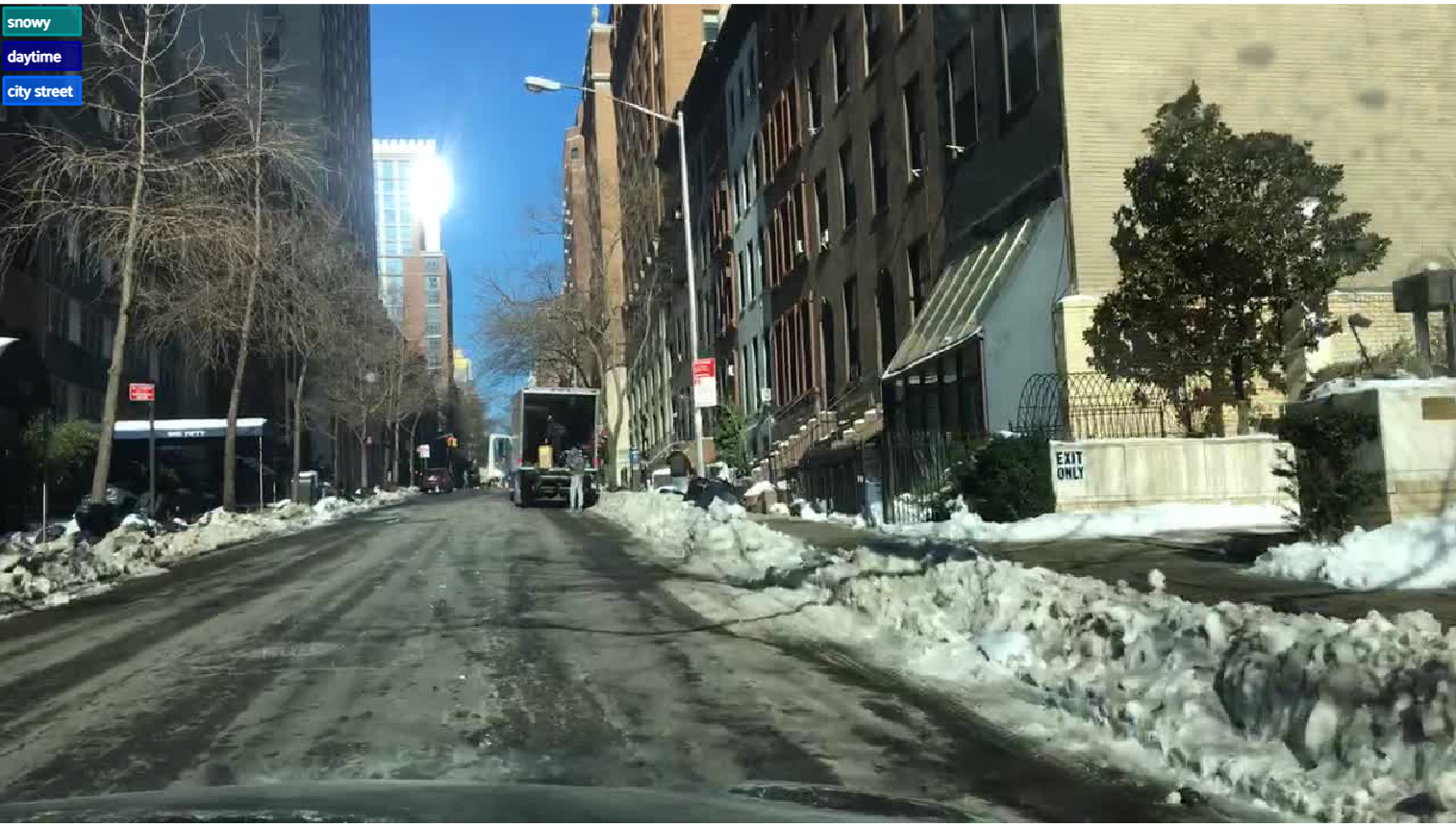}
		\caption{\textcolor{green}{$\mathcal{D}_9$}: snowy,city-street,daytime}
		\label{fig:d2cloud}
	\end{subfigure}
	
	\begin{subfigure}{.31\textwidth}
		\centering
		\includegraphics[width=\linewidth]{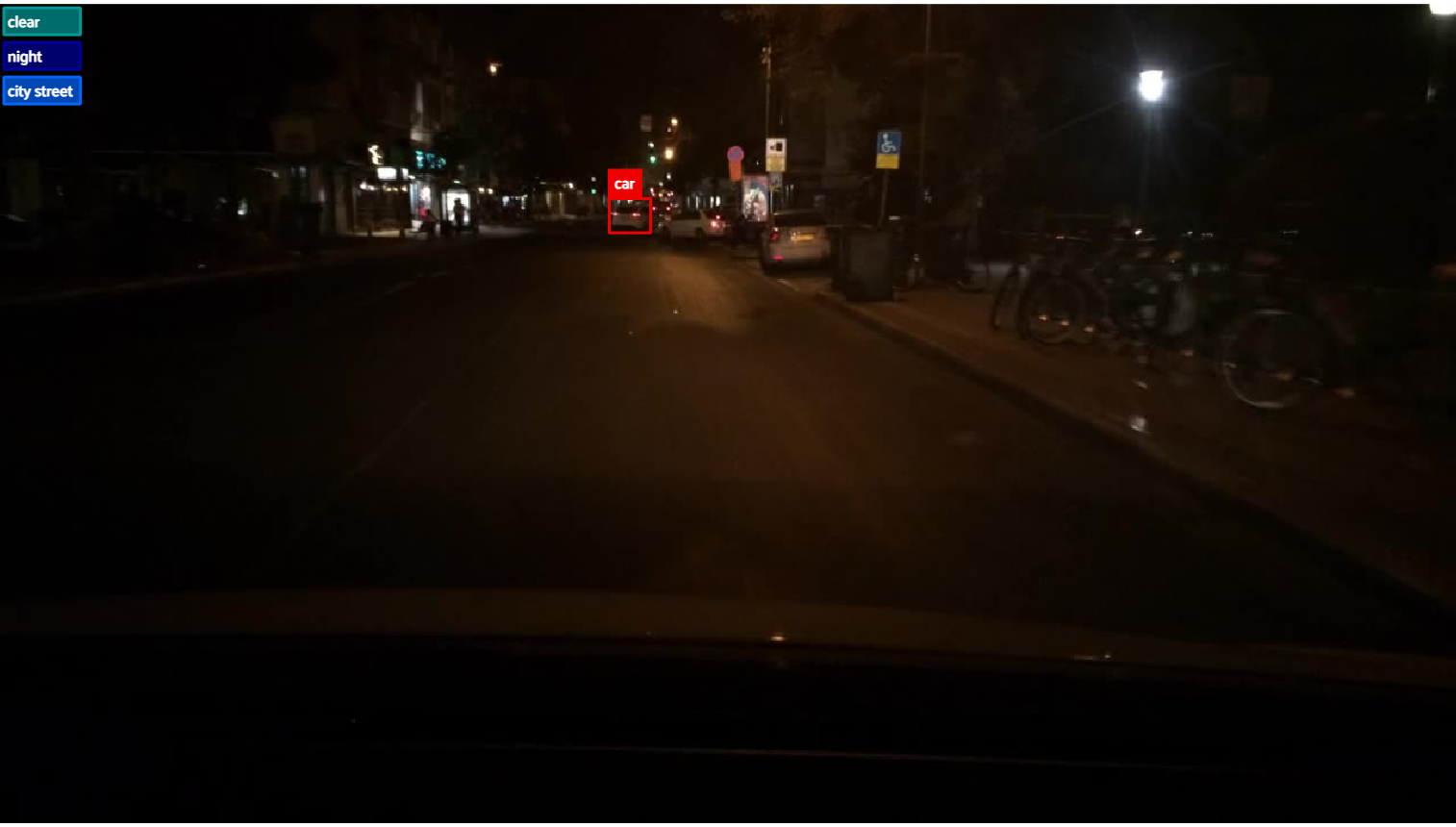}
		\caption{\textcolor{red}{$\mathcal{D}_1$}: clear,city-street,night}
		\label{fig:d3cloud}
	\end{subfigure}%
	\quad 
	\begin{subfigure}{.31\textwidth}
		\centering
		\includegraphics[width=\linewidth]{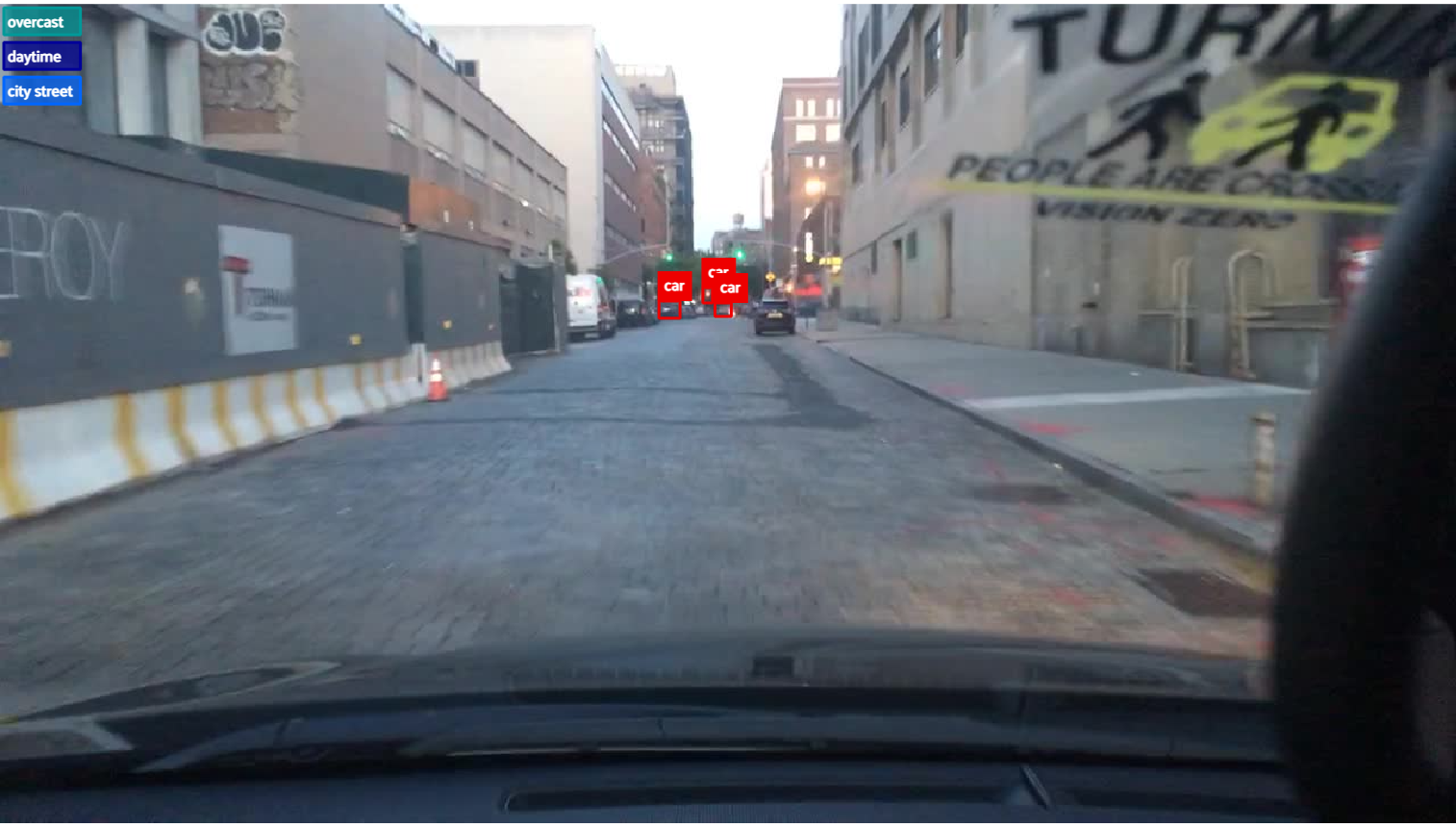}
		\caption{\textcolor{red}{$\mathcal{D}_7$}: overcast,city-street,daytime}
		\label{fig:d4cloud}
	\end{subfigure}
	
	\caption{Representative images of the discovered dataclouds}
	\label{fig:dataImgs}
\end{figure*}

\subsection{Experiment 1: Human-Interpretability}\label{sec:fuzzLearn}
We conduct an experiment to qualitatively evaluate the interpretability of the learned fuzzy monitor in this section. 
\subsubsection{Learned Fuzzy Monitor}
We trained a $\Xi$ from scratch using the $d_{train}$ dataset as described in section \ref{subsec:learncloud}. After training, the saved state of $\Xi$ yielded $10$ dataclouds, i.e. $\mathcal{D}_{0\dots 9}$. 

Representative images of the corresponding discovered prototypes for their respective dataclouds are shown in Fig.~\ref{fig:dataImgs}.     

\begin{figure}[htbp]
\centering
\begin{lstlisting}[
	language=ODD,
	backgroundcolor=\color{lightgrey},
	commentstyle=\color{codegreen},
	keywordstyle=\color{codeblue},
	numberstyle=\tiny\color{codegray},
	stringstyle=\color{codepurple},
	basicstyle=\ttfamily\scriptsize,
	breakatwhitespace=false,
	breaklines=true,
	captionpos=b,
	keepspaces=true,
	numbers=left,
	numbersep=2pt,
	showspaces=false,
	showstringspaces=false,
	showtabs=false,
	tabsize=2,
	]
Include weather is [clear, snowy,overcast,partly cloudy] 
Include scene is [city-street,highway]
Include timeofday is [daytime,dawn/dusk,night]
 
##Conditional ODD Statements
#Exclude Datacloud 1 
Conditional Exclude 
    [brightness,contrast_score,clearness_score] of visibility 
        for [clear, city-street, night] is [18.167,2.18,0.12]
## Exclude Datacloud 7
Conditional Exclude 
    [brightness,contrast_score,clearness_score] of visibility 
        for [overcast, city-street, daytime] is [111.2,3.284,0.48]
	
\end{lstlisting}
\caption{ISO34503 compliant ODD Specification}
\label{fig:oddspec}
\end{figure}

Additionally, we obtained 10 fuzzy rules (cf. \eqref{eq:fuzzRule}) which provide a mapping of the dataclouds to the yolov5x6 model behavior (e.g., \textit{reliable} ($\varphi=0$) or \textit{unreliable} ($\varphi=1$)). The discovered prototypes corresponding to dataclouds $\mathcal{D}_1$ and $\mathcal{D}_7$ map to unreliable behavior, see Fig. \ref{fig:d3cloud} and \ref{fig:d4cloud} containing misperceptions (\textit{red} bounding boxes); other dataclouds map to reliable behavior. 


\subsubsection{Human-Readable Specification}
The ISO34503~\cite{iso34503:2023} standard provides guidelines for creating a human-readable ODD specification. We employ the \textit{default} specification mode, requiring the implementation of ODD-exit monitor~\cite{on2021taxonomy} for \textit{included} and \textit{excluded} operating conditions mentioned in the ODD specification.

By shortlisting the dataclouds for which the yolov5x6 perception component exhibits reliable behavior (i.e., no mP), it is possible to create a human-readable ODD specification, as per ISO34503 guidelines. We treat prototypes shown in Fig~\ref{fig:dataImgs} as representative of the PO conditions; see section \ref{subsub:protopo}. Based on the mapping provided by the discovered fuzzy-rule (see Fig.~\ref{fig:dataImgs}, \textit{red}/\textit{green} text annotations), the operating conditions from $\mathcal{D}_{2}$, $\mathcal{D}_{6}$, $\mathcal{D}_{9}$ along with other shortlisted dataclouds (i.e., mapping to reliable yolov5x6 behavior), we can create an ODD specification as shown in Fig.~\ref{fig:oddspec}, see Line 1-3. 

However, the dataclouds $\mathcal{D}_1$ and $\mathcal{D}_7$ must be excluded from this ODD. This is accomplished by using Conditional ODD statements, as per ISO34503 guidelines, cf. Fig.~\ref{fig:oddspec}, Lines 5-14. The PO conditions corresponding to $\mathcal{D}_1$ comprise platform-specific operating conditions, collectively captured by intermediate operating condition \textit{visibility}, are shown in Fig.~\ref{fig:oddspec}, Lines 7-9. This conditional statement implies that the yolov5x6 perception component is prohibited from operating in \textit{clear} weather conditions on \textit{city streets} during the \textit{night} if the \textit{visibility} operating conditions belong to $\mathbf{p}_1$. Similarly, the conditional statement corresponding to the Conditional Exclude of $\mathcal{D}_7$ has been shown in Lines 11-14.  

The operating conditions mentioned in the ODD specification must be monitored, and the use of the perception component should be prohibited upon encountering excluded operating conditions. Authors in \cite{salvi2022fuzzy} provide a method for the fuzzy interpretation of ODD and its monitoring. However, the task of ODD monitoring is beyond the scope of this work. In the next section, we will discuss the use of discovered data clouds to gather evidence for creating an assurance case related to the safe operation of ADS using the yolov5x6 perception component within our specified ODD context.

\subsubsection{Assurance Case Creation}
\begin{figure*}
    \centering
    \includegraphics[width=0.8\textwidth]{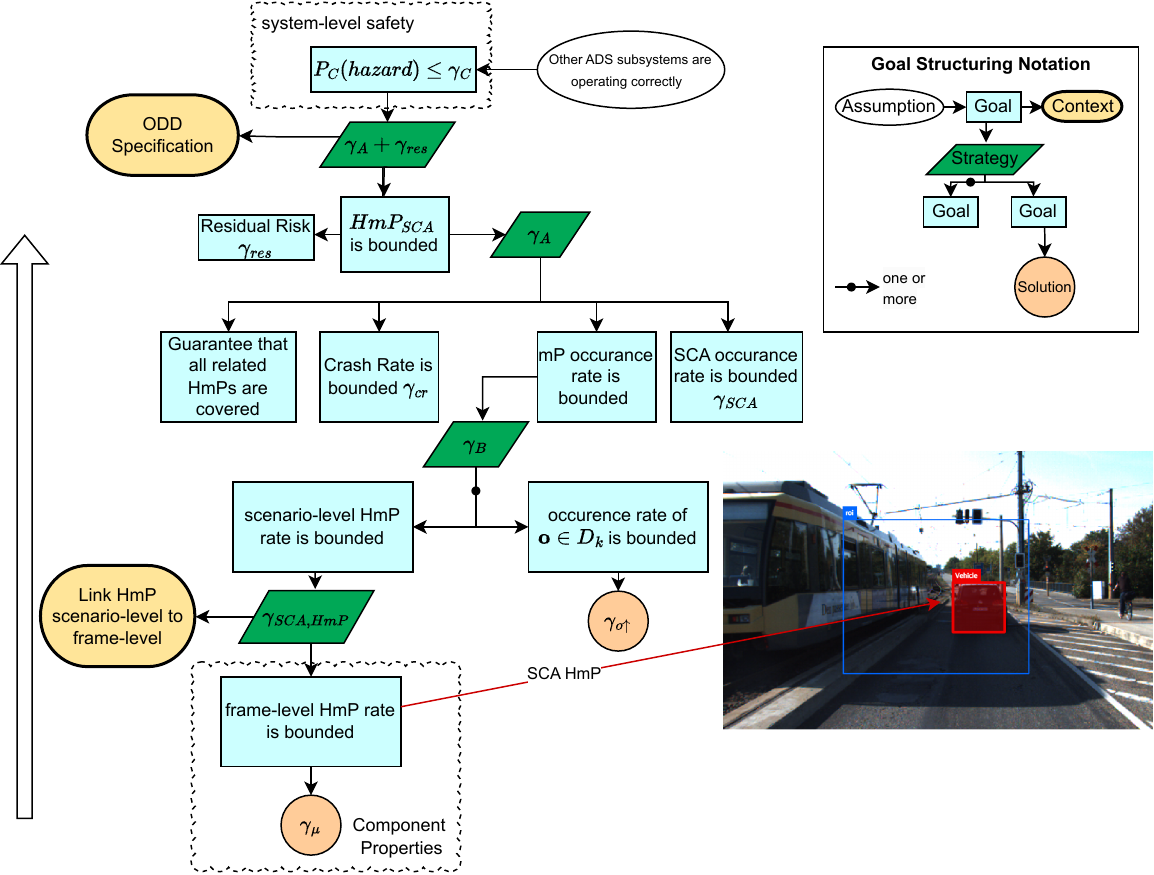}
    \caption{Safety Assurance Case for use-case scenario Stopped Car Ahead (SCA)}
    \label{fig:scE}
\end{figure*}

ISCaP establishes a connection between the top-level safety goal (i.e., the probability bound on misperception-caused collision $P_C(HmP)\leq\gamma_C$) and the bottom-level component properties (i.e., misperception (mP) occurrence rate). A simplified abstraction of the ISCaP safety assurance case for our SCA HmP is depicted in Fig.~\ref{fig:scE}. The bound on the top-level safety goal is initially broken down for our SCA HmP, as per the $\gamma_A$ decomposition. The decomposed sub-goals can be described as follows:
\begin{itemize}
    \item Guarantee that all related HmPs in the SCA use case are covered; we assume this is met for our evaluation.
    \item Conservatively assuming that all missed SCAs are hazardous, i.e., leading to rear-end collision.
    \item The occurrence rate of SCA is bounded, assuming SCA is encountered at every 500m distance.
    \item mP occurrence rate factoring different PO conditions within ODD context is bounded.
\end{itemize}


To create a safety assessment case for our use-case scenario, we instantiate the ISCaP template and obtain evidence for the leaf-level solutions (see Fig.~\ref{fig:scE}, orange \textit{Solution}). 
The evidence comprises of the frame-level HmP occurrence rate within a 
 set of PO conditions ($\gamma_{\mu}$) and occurrence probability of these PO conditions ($\gamma_{o\uparrow}$), and is calculated as\footnote{accounting for sampling error of $\Delta_{0.99}$}:
\begin{subequations}\label{eq:claims}
		\begin{equation}\label{eq:expo}
			\gamma_{o\uparrow} = \frac{|supp(\mathcal{D}_k)|}{|\Omega|}
		\end{equation}
		\begin{equation}\label{eq:hmp}
			\gamma_{\mu} =\frac{1}{|supp(\mathcal{D}_k)|}\sum_{\mathcal{D}_k}1[HmP]
		\end{equation}
\end{subequations}
As described in section \ref{subsub:evidence}, we collect the bottom-level evidence $\gamma_\mu$ and $\gamma_{o\uparrow}$ in order to perform a bottom-up computation of $\gamma_A$ corresponding to our target HmP concerning SCA; results are presented in Table~\ref{tab:SCVal}. The calculation of mP occurrence rate (cf. Fig.~\ref{fig:scE}, $\gamma_B$) in the safety case excludes evidence pertaining to $\mathcal{D}_1$ and $\mathcal{D}_7$ since they are excluded in our ODD specification (cf. Fig.~\ref{fig:oddspec}).   
\begin{table}[htbp]
	\centering
	\caption{Accumulated Evidence for ISCaP Safety Case}
	\begin{tabularx}{\columnwidth}{|l|X|c|} 
		\hline
		\textbf{Safety Bound} & \textbf{Description} & \textbf{Value for $\mathcal{D}_{0,2,3,4,5,6,8,9}$} \\
		\hline
		\textit{mP occurrence rate} & $\gamma_B = \sum_k (\gamma_{sca,HmP}\cdot\gamma_{o\uparrow}$) & $2.3185\times 10^{-1}$\\
		\hline
		\textit{SCA occurrence rate} & Based on SCA occurrence every 500m: $\gamma_{SCA}=\frac{40[km/hr]/10[fps]}{500[m]}$ & $2.2\times 10^{-3}$\\
		\hline
		\textit{crash rate}& Conservatively assume $HmP$ always leads to crash ($\gamma_{cr}=1$)& 1.0\\
		\hline
		\textit{$HmP_{SCA}$ rate} & $\gamma_A=\gamma_{cr}\cdot\gamma_B\cdot\gamma_{SCA}$ & $5.1 \times 10^{-4}$\\
		\hline
	\end{tabularx}	
	\label{tab:SCVal}
\end{table}

Given a saved training state of $\Xi$ (cf. section \ref{subsec:learncloud}), assuming a user-defined threshold $\gamma_C$, the calculated value of residual safety bound ($\gamma_{res}=\gamma_C-\gamma_A$) must be reviewed by a human safety-expert to assess the validity of created safety case. If judged unacceptable, the $\Xi$ training could be resumed with additional training data to discover new dataclouds, cf. Fig.~\ref{fig:proc}.  

The \textit{qualitative} evaluation results highlight the interpretability of our learned fuzzy monitor, demonstrated through the creation of human-readable ODD Specification (cf. Fig.~\ref{fig:oddspec}), supplemented with a safety assurance case (cf. Fig. \ref{fig:scE} and Table~\ref{tab:e1}).  

\subsection{Experiment 2: Runtime Safety Monitoring}\label{subsec:runEval}
We conduct a follow-up experiment to quantitatively evaluate the learned fuzzy monitor as a runtime safety monitor using the ML monitor evaluation framework provided by \cite{guerinUnifyingEvaluationMachine2022}. Unlike generic evaluation metrics such as accuracy, recall, and precision, which consider every mP equally hazardous, this framework provides flexibility to model the specificity of ML-based perception components.     

We assume the existence of an ODD-exit monitor for our specified ODD. Therefore, for this experiment, we filtered the validation dataset $d_{val}$ based on the ODD (cf. Fig.~\ref{fig:oddspec}). 87\% of the samples were found to be within this ODD.
 
\subsubsection{Evaluation Criteria} \label{subsec:escheme}
This framework measures the safety and reliability improvement achieved by the use of a runtime safety monitor ($m_C$) along with perception component $C$, i.e., $(C,m_C)$, based on the following criteria:
\begin{itemize}
	\item \textit{Safety Gain} ($SG_{m_C}$): safety improvements resulting from $m_C$,
	\item \textit{Residual Hazard} ($RH_{m_C}$): remaining hazard in the system after using $m_C$,
	\item \textit{Availability Cost} ($AC_{m_C}$): system performance decreases because of $m_C$.
\end{itemize}

\begin{subequations}
    \begin{equation}
        SG_{m_C}=\frac{1}{n}\sum_{i=1}^n \left( {R}^{\mathcal{S}}_{(C,m_C)} (x_i) - {R}^{\mathcal{S}}_{(C)} (x_i) \right)
    \end{equation}
    \begin{equation}
        RH_{m_C}=\frac{1}{n}\sum_{i=1}^n \left( {R}^{\mathcal{S}}_{C^*} (x_i) - {R}^{\mathcal{S}}_{(C,m_C)} (x_i) \right)
    \end{equation}
    \begin{equation}
        AC_{m_C}=\frac{1}{n}\sum_{i=1}^n \left( {R}^{\mathcal{M}}_{C} (x_i) - {R}^{\mathcal{M}}_{(C,m_C)} (x_i) \right)
    \end{equation}
\end{subequations}

These metrics are calculated based on the cumulative safety return $R^{\mathcal{S}}$ and mission return $R^{\mathcal{M}}$ over $n$ samples, assuming a good coverage of an ODD. The $R^{\mathcal{S}}_{C^*}$ represents the reward calculation assuming ground-truth knowledge ($y$) during $C$ usage. A binary variable $\tau$ represents the \textit{ground-truth} error status of $C$. We define the safety and mission returns as follows:

	\begin{equation}\label{eq:SR}
		{R}^{\mathcal{S}}_{(C,m_C)} (x)=
		\begin{cases}
			0 & \text{ if } \tau = 1 \text{ and } m_C=0\\
			1 & otherwise
		\end{cases}
	\end{equation}\break
	\begin{equation}\label{eq:MR}
		{R}^{\mathcal{M}}_{(C,m_C)} (x)=
		\begin{cases}
			0 & \text{ if } \tau = 0 \text{ and } m_C=1\\
			1 & otherwise
		\end{cases}
	\end{equation}
The monitored system ($C,m_C$) is penalized if the error exists, $\tau=1$), but the runtime monitor does not detect it, $m_C=0$, (i.e., \textit{fn}). Additionally, the system ($C,m_C$) must be penalized if it detects an error, $m_C=1$, but there is no error, $\tau=0$ (i.e., \textit{fp}). Please note the distinction between the incorrect prediction of the ML-based object detector (FN, FP) and monitor detections (\textit{fn}, \textit{fp}).

\subsubsection{Monitor Benchmarking Results}
We define two types of error status: mP status ($\tau_{mP}$) and HmP status ($\tau_{HmP}$). The $\tau_{mP}=1$ indicates a misperception (i.e., missed object), and $\tau_{HmP}=1$ indicates a hazardous misperception (i.e., collision with SCA due to mP).
We benchmark the performance of our proposed $m_C$, which uses the learned fuzzy monitor $\Xi$ as a runtime safety monitor. To compare our approach with others, we replace our $\Xi$ with several other well-known classifiers~\cite{scikit-learn}; we trained these classifiers on $d_{train}$. Additionally, we used a \textit{Random} monitor comprising a random value generator ($\in \{0,1\}$) for $\hat \varphi$, inspired by \cite{ferreira2021benchmarking}. Firstly, we evaluate the monitor's ability to detect untrustworthy PO conditions ( i.e., $\hat \varphi$ prediction accuracy). The evaluation results are shown in Table \ref{tab:e1}; the \textit{green} arrow indicates desired high or low scores, with \textbf{bold} values indicating the best scores. 

\begin{table}[htbp]
    \centering
\caption{Benchmarking runtime monitors: $\tau_{mP}$}
\label{tab:e1}
    \begin{tabular}{|c|c|c|c|} \hline 
       $\Xi$ in $m_C$  & \textbf{SG} \textcolor{green}{$\uparrow$} & \textbf{RH}\textcolor{green}{$\downarrow$} & \textbf{AC}\textcolor{green}{$\downarrow$}\\ \hline \hline
       \textit{Random} & \textbf{0.2018} & \textbf{0.1942} & 0.2357\\ \hline
       Gaussian Naive Bayes~\cite{scikit-learn_gaussiannb} & 0.1846 & 0.2902 & 0.2082 \\ \hline 
       Decision Tree~\cite{scikit-learn-dc}& 0.19345 & 0.22086 & 0.20639 \\ \hline 
 Neural Network~\cite{learn20171}
 & 0.03273 & 0.44216 & 0.04974\\\hline  
      \rowcolor{gray!30} Fuzzy-based (ours)& 0.0175 & 0.3968 &\textbf{0.01447} \\ \hline
    \end{tabular}   
\end{table}

Secondly, we evaluate the monitor's ability to intervene only in hazardous situations since not every misperception is hazardous, limited to the SCA use case. The results of this evaluation are presented in Table~\ref{tab:e2}. 

\begin{table}
    \centering
\caption{Benchmarking runtime monitors: $\tau_{HmP}$}
\label{tab:e2}
    \begin{tabular}{|c|c|c|c|} \hline 
       $\Xi$ in $m_C$  & \textbf{SG} \textcolor{green}{$\uparrow$} & \textbf{RH}\textcolor{green}{$\downarrow$} & \textbf{AC}\textcolor{green}{$\downarrow$}\\ \hline \hline
       \textit{Random} & 0.001523 & 0.00228 & 0.431073\\ \hline
       Gaussian Naive Bayes~\cite{scikit-learn_gaussiannb} & \textbf{0.00218} & 0.00349 & 0.3906 \\ \hline 
       Decision Tree~\cite{scikit-learn-dc}& 0.00114 & 0.00380 & 0.3956 \\ \hline 
 Neural Network~\cite{learn20171}
 & 0.0048 & \textbf{0.00087} & 0.7686\\\hline  
       \rowcolor{gray!30} Fuzzy-based (ours)& 0.0 & 0.00495 &\textbf{0.03198} \\ \hline
    \end{tabular}   
\end{table}

\subsubsection*{Discussion}
A \textit{good} runtime monitor should increase system safety without decreasing its availability. Ideally, a runtime monitor must only intervene when necessary, in our case, HmPs. However, unnecessary interventions (i.e., \textit{fp}'s) result in a reduced availability of perception components.    


In conducting the experiment with samples satisfying derived ODD (cf. Fig.~\ref{fig:oddspec}), it is expected that the proportion of HmPs from missed SCA should be limited by $\gamma_A$ (refer to Table \ref{tab:SCVal}). Ideally, the perception component should operate reliably within the confines of this ODD, obviating the need for a monitor $m_C$. However, since the ODD for this particular case study (cf. Fig.~\ref{fig:oddspec}) was created with the data clouds discovered using training data $d_{train}$. Therefore, in the event of incorrect detections by $m_C$ over $d_{val}$ (i.e., \textit{unseen} data during training), the runtime monitor $m_C$ has been implemented as an online fault tolerance mechanism, and its performance has been evaluated accordingly.

In case of the evaluation for the ability to accurately detect misperceptions (cf. Table \ref{tab:e1}), we see that Gaussian Naive Bayes and Decision Tree monitors are as good as a \textit{Random-monitor} when it comes to improving safety gains (SG) and reducing residual hazard (RH). This partially confirms the statement by \cite{ferreira2021benchmarking}, which states that existing ML input-checking monitors might be as good as random monitors. Regardless, we see from the high value of availability costs (AC) for these classifiers that the outputs from $C$ will not be utilized in driving functions due to a large number of \textit{fp}'s by $m_C$. This leads to frequent disengagement of the dependent driving functions, e.g., frequent driver-take-over requests. Neural Network-based classifier has comparatively lower SG but higher RH, indicating an inability to improve safety no better than other classifiers. However, it exhibits significantly lower AC, indicating fewer \textit{fp}'s.  In comparison, our fuzzy-based monitor shows very low AC scores and comparable RH scores to that of other classifiers. Since we already deploy $C$ in PO conditions where it operates reliably (i.e., ODD), we argue that the low SG is owing to limited potential for safety gains by introducing $m_C$. 

Within the context of our SCA use case, we have evaluated the ability of runtime safety monitors to prevent actual hazardous misperception(s) and the ability to ignore benign misperceptions. Gaussian Naive Bayes provides the highest SG scores, while the neural network-based classifiers exhibit the highest RH reduction. However, our fuzzy monitor has comparable RH reduction capability without incurring high AC like other monitors. This indicates the ability of our fuzzy-based monitor to safeguard the perception component against SCA HmP without impacting the availability of perception components, unlike other evaluated classifiers. 
Other evaluated classifiers were trained using batch data, with multiple training passes for neural network-based classifiers. In contrast, our fuzzy-based classifier was trained using single training instances incrementally in an online "test-then-train" manner, showing its prediction accuracy improvements during training.  

In summary, we have evaluated our proposed fuzzy monitor both qualitatively and quantitatively. The results highlight its human interpretability and comparable performance as a runtime safety monitor. Due to this interpretability and transparent structure of the monitor, for our automated driving case study, we were able to create a human-readable ODD specification (along with a safety assurance case). 

\section{Limitations and Threats to Validity}
Our approach utilizes a data-driven fuzzy monitoring learning technique, meaning that the accuracy of misperception detection significantly depends on data quality and training volume. The quality of human-provided labels within the dataset for operating condition annotations vastly impacts the quality of learned prototypes. Based on the AlMMo learning settings, a large number of dataclouds might be discovered, requiring increased human efforts for their review. 

The case study aims to provide human-interpretable sample prototypes for creating ODD specifications, required for human-driven activities like safety assurance cases and monitoring. However, using a single prototype for an ODD specification may lead to \textit{specification} uncertainty due to sampling error, negatively affecting its representativeness. We recommend using large labeled datasets for learning and further testing, supplemented by additional fault tolerance mechanisms to improve the derived ODD coverage and improve safety. The evaluation results assume an ODD-exit monitor for the monitoring task and to prevent operation in scenarios outside the ODD.

\section{Conclusion and Future Outlook}\label{sec:conc}
We have introduced our approach for learning a human-interpretable monitor for an ML-based perception component that is influenced by external operating conditions. The transparent structure of the learned fuzzy-based monitor enables the collection of evidence for the reliable operation of the perception component. We evaluated the learned fuzzy monitor using an automated driving case-study. Our qualitative evaluation results indicate that the learned fuzzy-monitor is well-suited for ODD specification (specifically, identifying a set of operating conditions where the perception component performs reliably) and can support human-driven Verification and Validation activities. The quantitative evaluation shows that when used alongside the perception component, the fuzzy-monitor can enhance \textit{safety} by reducing residual hazards (low RH) without compromising the perception component's \textit{availability} (low AC). Therefore, it can serve as an effective online fault tolerance mechanism. 
 
The learned monitor explains the ML model behavior impact in the form of human-interpretable fuzzy rules which are easy to understand. This understanding is essential for creating strong assurances for the reliable operation of the perception component and the safety of the system that utilizes the perception component. 
As future work, we plan to investigate the use of online fuzzy monitor learning in DevOps cycles for scaling its operation to diverse operating conditions through (semi-)automated ODD expansion of Automated Driving System features.

\bibliographystyle{unsrt}
\bibliography{references.bib}

\end{document}